\documentclass[11pt]{article}

\usepackage[preprint]{acl}

\usepackage{times}
\usepackage{latexsym}
\usepackage[T1]{fontenc}
\usepackage[utf8]{inputenc}
\usepackage{microtype}
\usepackage{graphicx}
\usepackage{amsmath}
\usepackage{amssymb}
\usepackage{booktabs}
\usepackage{multirow}
\usepackage[dvipsnames]{xcolor}
\usepackage{colortbl}
\usepackage{array}
\usepackage{adjustbox}
\usepackage{subcaption}
\usepackage[many]{tcolorbox}
\usepackage{pifont}
\usepackage{pgfplots}
\pgfplotsset{compat=1.18}

\definecolor{bestcolor}{HTML}{E8F5E9}
\newcommand{\best}[1]{\textbf{#1}}
\newcommand{\gain}[1]{\textcolor{green!50!black}{#1}}
\newcommand{\loss}[1]{\textcolor{red!70!black}{#1}}

\title{DuoMem: Towards Capable On-Device Memory Agents \\ via Dual-Space Distillation}

\author{
  \textbf{Peyman Hosseini\textsuperscript{1,2}} \Thanks{Work done during internship at Samsung R\&D Institute UK.} \quad
  \textbf{Ondrej Bohdal\textsuperscript{1}} \quad
  \textbf{Ahmed Alajrami\textsuperscript{1}} \quad
  \textbf{Andrea Maracani\textsuperscript{1}} \\
  \textbf{Ignacio Castro\textsuperscript{2}} \quad
  \textbf{Matthew Purver\textsuperscript{2}} \quad
  \textbf{Mete Ozay\textsuperscript{1}} \quad
  \textbf{Savas Ozkan\textsuperscript{1}} \quad
  \textbf{Taha Ceritli\textsuperscript{1}} \quad
  \\[0.5em]
  \textsuperscript{1}Samsung R\&D Institute UK \quad
  \textsuperscript{2}Queen Mary University of London \\
  \textbf{Correspondence}: \href{s.hosseini@qmul.ac.uk}{s.hosseini@qmul.ac.uk}, \href{o.bohdal.1@samsung.com}{o.bohdal.1@samsung.com} \\
}

\begin{document}
\maketitle

\begin{abstract}
Large language model (LLM)-based agents can solve complex procedural tasks by interacting with environments over multiple turns, but this ability typically depends on large models, long contexts, and repeated inference calls.
This makes advanced memory-augmented agents difficult to deploy on resource-constrained devices.
We introduce \textbf{DuoMem}, a dual-space distillation framework that transfers procedural problem-solving ability from a large teacher model to compact student models.
DuoMem distils in two complementary spaces: (1)~\emph{context-space distillation}, which replaces student-generated memories with higher-quality teacher-generated procedural memories prepended to the student's input, and (2)~\emph{parameter-space distillation}, which fine-tunes lightweight LoRA adapters on successful teacher trajectories.
Evaluated on ALFWorld, a challenging embodied decision-making benchmark, DuoMem boosts a 4B-parameter model from 4.3\% to 77.9\% task success rate, closing most of the gap to a 72B teacher model (87.1\%), while adding fewer than 10M trainable parameters and only a few megabytes of pre-computed teacher memories. Moreover, the DuoMem-enhanced 4B model completes tasks over 3$\times$ faster than the 72B teacher in wall-clock time, making it viable for real-time edge deployment, which would be challenging for the teacher.
Extensive ablations across eight models spanning 2B--72B parameters reveal that both distillation axes contribute complementary gains, with the combination yielding improvements far exceeding either component in isolation.
\end{abstract}

\section{Introduction}
\label{sec:intro}

Recent advances in large language models (LLMs) have enabled impressive reasoning and problem-solving capabilities \citep{wei2022chain, brown2020language}. When deployed as interactive agents, LLMs can engage in multi-turn exchanges with environments, iteratively planning and executing actions to accomplish complex tasks \citep{yao2022react, wang2024agent}.
A crucial ingredient for such agents is \emph{procedural memory}, defined as the ability to store and reuse structured knowledge derived from past interaction trajectories \citep{sumers2023cognitive, park2023generative}. By summarizing previously successful strategies into reusable scripts or condensed memory entries, agents can bootstrap performance on novel but structurally related tasks \citep{wang2024voyager, majumder2024clin, kagaya2024rap}.

\begin{figure}[t]
    \centering
    \includegraphics[width=0.4\textwidth]{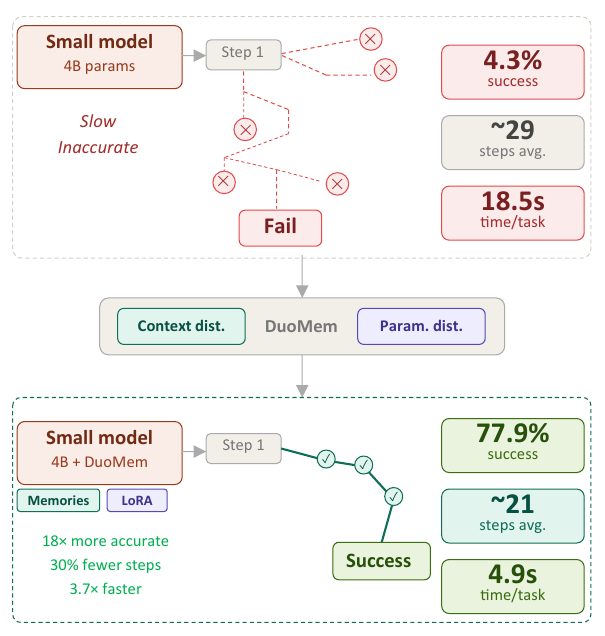}
    \caption{With \textbf{DuoMem}, a compact 4B model overcomes the procedural reasoning limitations of small LLMs. \emph{Top:} without DuoMem, the agent explores many wrong paths and fails (4.3\% success, $\sim$29 steps avg.). \emph{Bottom:} after dual-space distillation, the same model follows efficient trajectories and succeeds (77.9\% success, $\sim$21 steps avg.), approaching the 72B teacher (87.1\%).}
    \label{fig:method_overview}
    \vspace{-1.5em}
\end{figure}

Despite these advances, the computational cost of capable LLM agents remains a significant barrier. State-of-the-art performance typically requires models with 30B--70B+ parameters and long multi-turn context windows, consuming substantial memory, energy, and latency that prohibit deployment on edge devices such as smartphones, robots, or embedded systems \citep{wang2024agent}. Smaller models (2B--12B parameters), while attractive for on-device use, exhibit significantly weaker procedural reasoning and often fail to leverage memory effectively even when provided.

We address this gap with \textbf{DuoMem} (Figure~\ref{fig:method_overview}), a framework that distils procedural memory capabilities from a large teacher model into compact student models through two complementary mechanisms operating in different ``spaces'':

\begin{enumerate}
\item \textbf{Context-Space Distillation (CD):} Rather than relying on the student model to generate its own memories from past trajectories, we use a powerful teacher model to extract higher-quality procedural memories offline. At inference, most relevant teacher-generated memories are prepended to the student's context, enriching it with distilled procedural knowledge without modifying any model parameters.

\item \textbf{Parameter-Space Distillation (LoRA):} We collect successful teacher trajectories on training tasks and use them to fine-tune lightweight Low-Rank Adaptation (LoRA) modules \citep{hu2022lora} on the student model, teaching it to better follow procedural patterns observed in expert demonstrations.
\end{enumerate}

We evaluate DuoMem on ALFWorld \citep{shridhar2021alfworld}, a challenging text-based embodied benchmark requiring agents to complete household tasks via multi-step planning and interaction. Our experiments span 8 models from 2B to 72B parameters across, Qwen and Gemma families.

\paragraph{Contributions.}
\textbf{(i)}~We propose DuoMem, a dual-space distillation framework that transfers procedural memory from a large teacher to compact students via complementary context-space (CD) and parameter-space (LoRA) distillation.
\textbf{(ii)}~On ALFWorld, DuoMem lifts a 4B model from 4.3\% to 77.9\% task success, closing 89\% of the gap to the 72B teacher, while completing tasks over 3$\times$ faster, with fewer than 10M added parameters and only a few megabytes of pre-computed memories.
\textbf{(iii)}~Ablations across eight models show that both distillation spaces contribute complementary gains whose combination consistently exceeds either in isolation, and that thinking-mode models, despite higher accuracy, incur 5--7$\times$ latency penalties that make non-thinking students with DuoMem more practical for edge deployment (Appendix~\ref{app:thinking}).

\section{Related Work}
\label{sec:related}

\paragraph{Procedural Memory for LLM Agents.}
Augmenting LLM agents with memory has emerged as a key strategy for multi-step task completion, with approaches spanning skill libraries \citep{wang2024voyager}, memory streams for social simulation \citep{park2023generative}, continual memory accumulation \citep{majumder2024clin}, retrieval-augmented planning \citep{kagaya2024rap}, and knowledge-constrained action planning \citep{zhu2025knowagent}.
Most closely related to our work, \citet{fang2025memp} proposed MemP, a task-agnostic framework that treats procedural memory as a first-class optimization object, studying build, retrieval, and update strategies for distilling trajectories into reusable scripts; we adopt MemP's memory architecture as the foundation of DuoMem.
While these works demonstrate the value of memory, they focus on large models and do not address how to transfer memory-based reasoning to smaller models suitable for on-device deployment. DuoMem bridges this gap via dual-space distillation. We discuss additional related work, including recent approaches such as AWM \citep{wang2025awm} and LEGOMem \citep{han2026legomem}, in Appendix~\ref{app:extended_related}.

\paragraph{Knowledge Distillation for LLMs.}
Knowledge distillation \citep{hinton2015distilling} transfers capabilities from a larger teacher to a smaller student model. Recent work has explored distillation for LLMs in various settings: \citet{hsieh2023distilling} showed that distilling chain-of-thought rationales can outperform larger models with less data, while \citet{fu2023distillation} demonstrated specialization of smaller models for multi-step reasoning.
\citet{xu2024survey} provide a comprehensive survey of LLM distillation approaches.
Our work differs by focusing specifically on procedural memory distillation in an agent setting, operating simultaneously in both context space and parameter space.

\paragraph{Parameter-Efficient Fine-Tuning.}
LoRA \citep{hu2022lora} and its variants such as QLoRA \citep{dettmers2023qlora} enable efficient adaptation by training only a small number of low-rank parameters while freezing the base model. We leverage LoRA as the parameter-space component of our distillation framework, training on successful teacher trajectories to transfer behavioral patterns.

\section{Method}
\label{sec:method}

\begin{figure*}[t]
    \centering
    \includegraphics[width=0.75\textwidth]{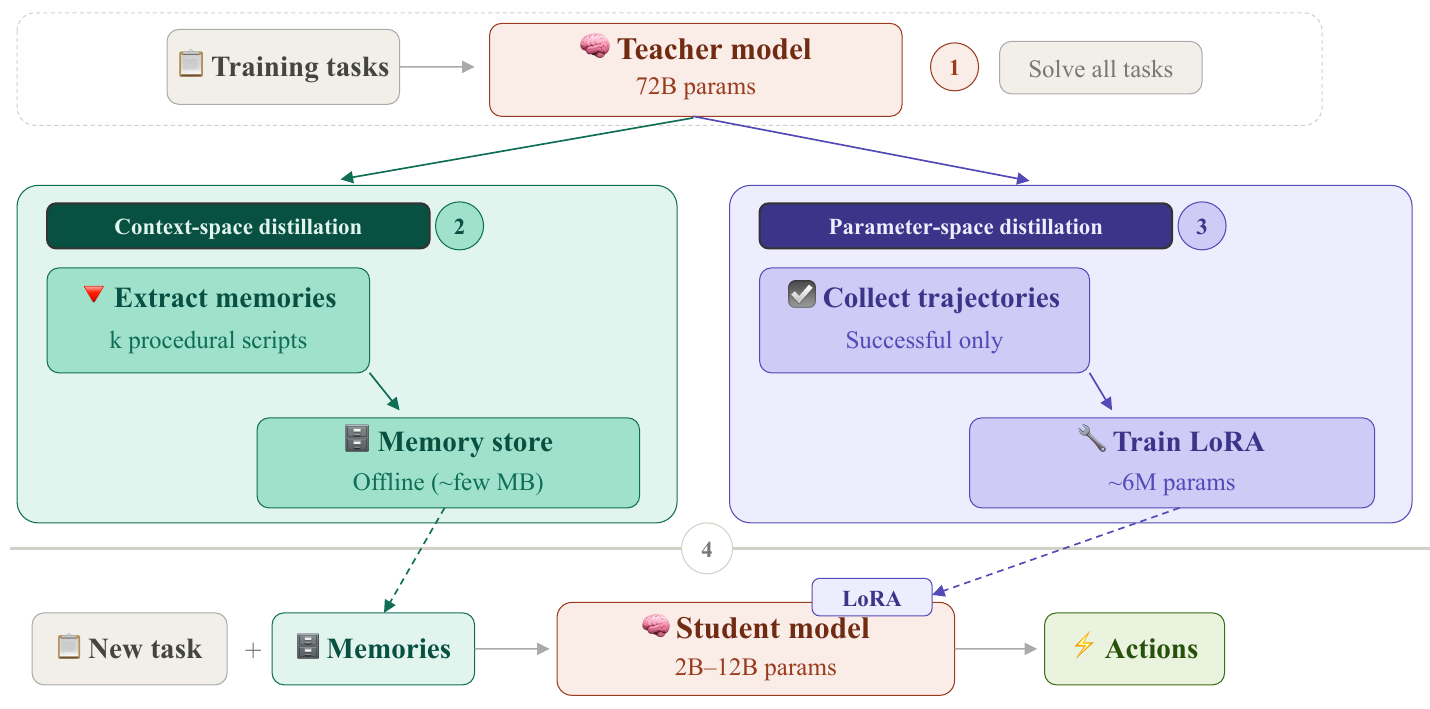}
    \caption{Overview of the \textbf{DuoMem} framework. \emph{Offline} (steps 1--3): the 72B teacher solves training tasks, producing (left) procedural memory scripts stored in a memory bank and (right) successful trajectories used to train LoRA adapters. \emph{Inference} (step~4): given a new task, the $k$ most relevant memories are retrieved via cosine similarity of semantic embeddings and prepended to the student's prompt; the LoRA-adapted student then executes the task.}
    \label{fig:method}
    \vspace{-1em}
\end{figure*}

We now describe DuoMem in detail, providing an overview in Figure~\ref{fig:method}. We first formalize the procedural memory framework (\S\ref{ssec:memp}), then present our two distillation mechanisms: context-space distillation (\S\ref{ssec:cd}) and parameter-space distillation (\S\ref{ssec:lora}), and finally describe their combination (\S\ref{ssec:duomem}).

\subsection{Procedural Memory for Agents}
\label{ssec:memp}

We consider an agent $\mathcal{A}_\theta$ parameterized by an LLM with parameters $\theta$ that interacts with an environment $\mathcal{E}$ to complete tasks. At each timestep $t$, the agent receives an observation $o_t$ and produces an action $a_t$. A complete interaction forms a \textbf{trajectory} $\tau = (o_1, a_1, o_2, a_2, \ldots, o_T, a_T)$, which ends either in task success or failure.

\textbf{Procedural memory} extends this framework by maintaining a memory store $\mathcal{M}$ that accumulates structured knowledge from past trajectories \citep{fang2025memp}. Given a set of $k$ completed trajectories $\{\tau_1, \ldots, \tau_k\}$ for tasks similar to a new target task, the agent extracts memories using a \emph{memory extraction function}:
\begin{equation}
m_i = f_\text{extract}(\tau_i; \theta), \quad i = 1, \ldots, k
\label{eq:extract}
\end{equation}
where $f_\text{extract}$ prompts the LLM to summarize each trajectory into a concise procedural script. For a new task with description $d$ and environment state $e$, the agent's input becomes:
\begin{equation}
x = [s;\; \tau_\text{ref};\; e;\; d;\; m_1; \ldots; m_k]
\label{eq:input}
\end{equation}
where $s$ is the system prompt (defines available actions and output format), $\tau_\text{ref}$ is a single full reference trajectory provided as a few-shot demonstration, and $m_1, \ldots, m_k$ are the retrieved memory scripts appended to the task observation. The system prompt is given as a system-role message, the reference trajectory as alternating user/assistant turns, and the observation---consisting of $e$, $d$, and the memory scripts---as the initial user message (see Appendix~\ref{app:qualitative}, Figure~\ref{fig:pipeline_anatomy} for a detailed illustration). $[\cdot;\cdot]$ denotes concatenation.

We study three memory configurations of increasing richness:
\begin{itemize}
\item \textbf{Script}: Provides $k$ concise procedural scripts summarizing past trajectories.
\item \textbf{Trajectory}: Provides a single full reference trajectory $\tau_\text{ref}$ (without script summaries) -- hand-crafted to be more useful.
\item \textbf{MemP (Full Procedural Memory)}: Provides both $k$ scripts and one full trajectory, representing the richest memory configuration.
\end{itemize}

\subsection{Context-Space Distillation}
\label{ssec:cd}

In the standard procedural memory setup, the agent using the student model $\mathcal{A}_{\theta_S}$ generates its own memory summaries from trajectories. However, smaller models often produce lower-quality summaries that miss key procedural steps or include irrelevant details.

Context-space distillation addresses this by replacing the student's memory extraction with the teacher's. Given a teacher model $\mathcal{A}_{\theta_T}$ with superior summarization ability, we use teacher memories for the most relevant $k$ completed trajectories:
\begin{equation}
m_i^T = f_\text{extract}(\tau_i; \theta_T), \quad i = 1, \ldots, k
\label{eq:cd}
\end{equation}

At inference, the student's input becomes:
\begin{equation}
x_\text{CD} = [s;\; \tau_\text{ref};\; e;\; d;\; m_1^T; \ldots; m_k^T]
\label{eq:cd_input}
\end{equation}
This is a \emph{training-free} distillation mechanism: we simply swap in the teacher's memories without modifying the student's parameters. The teacher memories are generated once offline and stored as text files (typically a few megabytes for the entire task suite), making this approach highly practical. Concretely, the teacher first generates memories for all completed trajectories offline. At inference, given a new task description $d$, we retrieve the $k$ most relevant memories by ranking all stored entries according to the cosine similarity of their semantic embeddings (computed with OpenAI's \texttt{text-embedding-3-small}) against the embedding of~$d$.

\subsection{Parameter-Space Distillation}
\label{ssec:lora}

While CD enriches the student's input context, parameter-space distillation directly improves the student's behavioral policy by fine-tuning on expert demonstrations.

\paragraph{Teacher Trajectory Collection.} We prompt the teacher model $\mathcal{A}_{\theta_T}$ to solve all training tasks in $\mathcal{E}$, retrying up to 5 times if a task has not been solved. We retain only \textbf{successful trajectories}, as training on failures would teach undesirable behavioral patterns. We denote the resulting dataset as $\mathcal{D} = \{(\hat{x}_j, \hat{y}_j)\}_{j=1}^{N}$, where $\hat{x}_j$ is the input context for trajectory $j$ and $\hat{y}_j$ is the corresponding successful action sequence.

\paragraph{LoRA Fine-Tuning.} We attach low-rank adapter matrices to the student model and train them via next-token prediction on the teacher's successful trajectories:
\begin{equation}
\mathcal{L} = -\sum_{j=1}^{N} \sum_{t=1}^{|\hat{y}_j|} \log p_{\theta_S + \Delta\theta}(\hat{y}_{j,t} \mid \hat{x}_j, \hat{y}_{j,<t}),
\label{eq:lora}
\end{equation}
where $\Delta\theta$ represents the LoRA parameters with rank $r$ and scaling factor $\alpha$. Only $\Delta\theta$ is updated during training; the base student model parameters $\theta_S$ remain frozen.

Concretely, each successful teacher trajectory $\tau^T = (o_1, a_1, \ldots, o_T, a_T)$ yields $T$ training samples. For the $i$-th sample, the input is the concatenation of the system prompt~$s$, the environment description~$e$, and the interaction history up to observation~$o_i$, i.e., $\hat{x}_i = [s; e; o_1, a_1, \ldots, o_{i-1}, a_{i-1}, o_i]$, and the target output is the teacher's action~$a_i$. The model is thus trained to imitate the teacher's policy at every decision point given the full preceding context.

\paragraph{Input Strategies.} We investigate three strategies for constructing the training input $\hat{x}_j$ from teacher trajectories: \textbf{\emph{(i)~Full History}}, where the entire multi-turn interaction history is provided and the model is trained to predict all assistant turns; \textbf{\emph{(ii)~Latest Only}}, where only the most recent turn serves as the training signal; and \textbf{\emph{(iii)~Last-5}}, where the five most recent turns are provided.

\subsection{DuoMem: Combining Both Spaces}
\label{ssec:duomem}

DuoMem applies both distillation mechanisms simultaneously. The student model is first LoRA-fine-tuned on teacher trajectories (parameter space), then at inference receives teacher-generated memories in its context (context space):
\begin{equation}
a_t = \mathcal{A}_{\theta_S + \Delta\theta}(x_\text{CD}, o_{\leq t}, a_{<t})
\label{eq:duomem}
\end{equation}

The two mechanisms are complementary: CD improves \emph{what} the model sees (richer procedural knowledge in context), while LoRA improves \emph{how} the model acts (better utilization of context for action generation). As we demonstrate empirically, their combination yields gains that substantially exceed either component alone.

\section{Experimental Setup}
\label{sec:setup}

\paragraph{Environment.}
We evaluate on \textbf{ALFWorld} \citep{shridhar2021alfworld}, a text-based embodied environment derived from the ALFRED benchmark \citep{shridhar2020alfred}.
ALFWorld contains six task types (e.g., picking up objects, heating/cooling items, examining objects) that require multi-step planning and interaction. Each task provides the agent with a textual description of the goal and environment observations. The agent must issue text-based actions (e.g., \texttt{go to countertop 1}, \texttt{pick up apple 1}) to navigate and manipulate the environment. We use the standard development split (140 tasks) for validation/hyperparameter selection and the test, which is labelled unseen validation in the environment, split (134 tasks) for final evaluation. For training, we use the standard 3,553 training tasks.

\paragraph{Models.}
We evaluate models from the Gemma \citep{google2025gemma, google2026gemma} and Qwen \citep{qwen2025qwen3} families spanning 2B--72B parameters: Gemma4-E2B-it, Qwen3-4B-Instruct, Qwen3-4B-Thinking, Qwen3-8B, Gemma3-12B-it, Qwen3-14B (in thinking and non-thinking modes), Qwen3-30B-A3B (MoE, 3B active), Qwen3-32B in both modes, and Qwen2.5-72B-Instruct as the teacher.
\paragraph{Teacher.}
We use \textbf{Qwen2.5-72B-Instruct} as the teacher model for both distillation mechanisms. For trajectory collection, we prompt the teacher to solve each of the 3,553 unique training tasks multiple times (3--4 attempts per task), yielding 11,546 task instances in total; this deliberate oversampling produces diverse solution paths for the same task, which enriches the training signal and helps reduce overfitting to a single trajectory per task. The teacher achieves a 99.0\% cumulative success rate across up to 5 retry attempts per instance, producing 11,434 successful trajectories. For CD, the teacher generates memory scripts that are stored offline and prepended to student inputs.

\paragraph{LoRA Training Details.}
We perform a hyperparameter search over LoRA rank $r \in \{8, 16, 32\}$ and learning rate $\text{lr} \in \{10^{-6}, 5 \times 10^{-6}, 10^{-5}, 5 \times 10^{-5}\}$, with $\alpha/r = 2$ for all configurations. Models are trained on the full training split using the \emph{Full History} input strategy (found to be best; see \S\ref{ssec:input_strategy}). We select the best checkpoint based on the MemP (W/ CD) success rate on the validation set. 

\paragraph{Metrics.}
We report \textbf{success rate} (\%): the percentage of tasks solved within the maximum allowed steps, and \textbf{average steps} (\#Steps) and \textbf{runtime} (Seconds): respectively the mean number of interaction rounds and time the model uses per task.

\section{Results}
\label{sec:results}

We organize our results to first present the full DuoMem performance (\S\ref{ssec:main}) and efficiency analysis (\S\ref{ssec:efficiency}), then analyze each component through ablations (\S\ref{ssec:ablation_cd}--\S\ref{ssec:ablation_nummem}).

\subsection{Main Results: DuoMem}
\label{ssec:main}

Table~\ref{tab:main} presents the main DuoMem results for models where both LoRA and CD have been applied, compared against baselines.

\begin{table*}[t]
\centering
\small
\begin{tabular}{@{}l c c c c c c c@{}}
\toprule
\textbf{Model} & \textbf{No Mem.} & \textbf{MemP} & \textbf{+CD} & \textbf{+LoRA} & \textbf{+DuoMem} & \textbf{$\Delta$ vs.\ No Mem.} & \textbf{$\Delta$ vs.\ MemP} \\
\midrule
Gemma4-E2B-it (2B) & 2.1 & 14.3 & 16.4 & 46.4 & \best{55.7} & \gain{+53.6 (+2552\%)} & \gain{+41.4 (+290\%)} \\
Qwen3-4B-Inst.\ (4B) & 4.3 & 55.0 & 56.4 & 72.1 & \best{77.9} & \gain{+73.6 (+1712\%)} & \gain{+22.9 (+42\%)} \\
Qwen3-8B (8B) & 40.7 & 64.3 & 61.4 & 60.7 & \best{64.3} & \gain{+23.6 (+58\%)} & \gain{+0.0 (+0\%)} \\
Gemma3-12B-it (12B) & 14.3 & 41.4 & 46.4 & 56.4 & \best{66.4} & \gain{+52.1 (+364\%)} & \gain{+25.0 (+60\%)} \\
\midrule
Qwen2.5-72B (Teacher) & 87.1 & 91.4 & --- & --- & --- & --- & --- \\
\bottomrule
\end{tabular}
\caption{Main DuoMem results on the test set (success rate \%). \textbf{MemP}: full procedural memory with student-generated memories. \textbf{+CD}: MemP with context distillation (teacher memories). \textbf{+LoRA}: MemP with parameter distillation (LoRA fine-tuning). \textbf{+DuoMem}: both CD and LoRA. $\Delta$~columns report the absolute and relative gain of DuoMem over No Memory and over MemP, respectively. DuoMem yields the strongest configuration for every student model, with the 4B model reaching 77.9\%, closing 89\% of the gap to the 72B teacher.}
\label{tab:main}
\end{table*}

The results demonstrate that DuoMem produces transformative improvements for small models. The most striking case is Qwen3-4B-Instruct, which improves from 4.3\% (no memory) to 77.9\% with DuoMem, an 18$\times$ relative improvement that closes 89\% of the absolute gap to the 72B teacher's 87.1\%. The 2B Gemma4-E2B-it model also benefits strongly from the full dual-space configuration, rising from 2.1\% without memory to 55.7\% with DuoMem, while Gemma3-12B-it rises from 14.3\% to 66.4\%. Together, these results show that DuoMem can substantially improve procedural problem solving across compact 2B--12B students.

Crucially, the decomposition into +CD and +LoRA columns reveals that both distillation spaces contribute, and their combination is superior.
For Gemma4-E2B-it, MemP alone gives 14.3\%, CD raises it to 16.4\%, and LoRA with MemP reaches 46.4\%; applying both distillation spaces yields 55.7\%, adding 9.3 points over LoRA+MemP and 39.3 points over CD alone.
For Qwen3-4B-Instruct, MemP alone gives 55.0\%; CD adds a modest 1.4 points (to 56.4\%), LoRA adds a substantial 17.1 points (to 72.1\%), and combining both in DuoMem yields 77.9\%, a further 5.8 points beyond MemP+LoRA, confirming that teacher-quality memories complement LoRA-improved action generation.
For Gemma3-12B-it, the pattern is even more pronounced: CD and LoRA each contribute independently (+5.0 and +15.0 points respectively), while DuoMem achieves +25.0 over base MemP, which is more than the sum of the individual gains, and suggests a synergistic interaction between the two distillation spaces.

As a preliminary finding motivating our work, we observe that procedural memory (MemP) yields large gains across all model sizes (see Appendix~\ref{app:full_results} for the full breakdown by memory type): the 4B Qwen3-4B-Instruct model improves by +1182\% (Table~\ref{tab:full_inference}) with MemP, while the 72B teacher improves by only +5\%, confirming smaller models benefit disproportionately from structured external memory. Thus, we adopt MemP's memory configuration for subsequent DuoMem experiments.

\subsection{Efficiency and Deployment Analysis}
\label{ssec:efficiency}

A key motivation for DuoMem is enabling on-device deployment, where both accuracy \emph{and} inference cost matter. Table~\ref{tab:efficiency} reports the average number of interaction steps and wall-clock task completion time across configurations for each student model and the 72B teacher, alongside the number of additional LoRA parameters.

\begin{table*}[t]
\centering
\small
\begin{adjustbox}{max width=\textwidth}
\begin{tabular}{@{}l l r r r r r r r r@{}}
\toprule
& & \multicolumn{5}{c}{\textbf{Configuration}} & \multicolumn{2}{c}{\textbf{DuoMem Gain}} & \\
\cmidrule(lr){3-7} \cmidrule(lr){8-9}
\textbf{Model} & \textbf{Metric} & \textbf{No Mem.} & \textbf{MemP} & \textbf{+CD} & \textbf{+LoRA} & \textbf{+DuoMem} & \textbf{$\Delta$ vs.\ No Mem.} & \textbf{$\Delta$ vs.\ MemP} & \textbf{LoRA Params} \\
\midrule
\multirow{2}{*}{Gemma4-E2B-it (2B)} & Avg.\ Steps ($\downarrow$) & 29.5 & 33.5 & 32.9 & 28.1 & \best{25.7} & \gain{$-$13\%} & \gain{$-$23\%} & \multirow{2}{*}{15.3M} \\
 & Avg.\ Time (s) ($\downarrow$) & 17.24 & 15.20 & 14.91 & 15.06 & \best{13.08} & \gain{$-$24\%} & \gain{$-$14\%} & \\
\midrule
\multirow{2}{*}{Qwen3-4B-Inst.\ (4B)} & Avg.\ Steps ($\downarrow$) & 29.3 & 24.8 & 24.3 & 21.9 & \best{20.6} & \gain{$-$30\%} & \gain{$-$17\%} & \multirow{2}{*}{5.9M} \\
 & Avg.\ Time (s) ($\downarrow$) & 18.49 & 14.07 & 11.16 & 5.89 & \best{4.89} & \gain{$-$74\%} & \gain{$-$65\%} & \\
\midrule
\multirow{2}{*}{Qwen3-8B (8B)} & Avg.\ Steps ($\downarrow$) & \best{23.7} & 23.8 & 24.3 & 24.4 & 24.2 & \loss{+2\%} & \loss{+2\%} & \multirow{2}{*}{18.9M} \\
 & Avg.\ Time (s) ($\downarrow$) & 14.75 & 12.82 & 13.27 & 10.25 & \best{9.98} & \gain{$-$32\%} & \gain{$-$22\%} & \\
\midrule
\multirow{2}{*}{Gemma3-12B-it (12B)} & Avg.\ Steps ($\downarrow$) & 27.2 & 27.4 & 26.2 & 24.3 & \best{23.2} & \gain{$-$15\%} & \gain{$-$15\%} & \multirow{2}{*}{23.6M} \\
 & Avg.\ Time (s) ($\downarrow$) & 23.21 & 15.16 & 15.42 & 15.04 & \best{13.81} & \gain{$-$40\%} & \gain{$-$9\%} & \\
\midrule
\multirow{2}{*}{Qwen2.5-72B (Teacher)} & Avg.\ Steps ($\downarrow$) & 14.7 & 17.6 & --- & --- & --- & --- & --- & \multirow{2}{*}{---} \\
 & Avg.\ Time (s) ($\downarrow$) & 20.57 & 16.69 & --- & --- & --- & --- & --- & \\
\bottomrule
\end{tabular}
\end{adjustbox}
\caption{Efficiency metrics across DuoMem configurations on the test set. \textbf{Avg.\ Steps}: mean interaction steps per task. \textbf{Avg.\ Time}: mean wall-clock time per task in seconds. \textbf{$\Delta$~columns}: relative change of DuoMem versus No Memory and MemP baselines (negative = improvement). \textbf{LoRA Params}: additional trainable parameters. The bottom row reports the 72B teacher model, which achieves fewer steps but requires 3--4$\times$ longer wall-clock time per task than the DuoMem-enhanced 4B student, highlighting the impracticality of deploying the teacher on resource-constrained devices.}
\label{tab:efficiency}
\vspace{-1em}
\end{table*}

The teacher model (Qwen2.5-72B-Instruct), despite using fewer interaction steps, requires an average of 16.7--20.6 seconds per task depending on the memory configuration. In contrast, the DuoMem-enhanced Qwen3-4B-Instruct completes tasks in just 4.89 seconds on average, a 3.4$\times$ speedup over the teacher's best configuration, while achieving a competitive 77.9\% success rate. This result underscores a key practical advantage: deploying large models on edge devices is infeasible not only due to memory requirements but also due to prohibitive inference latency. DuoMem enables a compact 4B model to deliver strong procedural performance at a fraction of the computational cost.

\paragraph{Parameter Overhead.}
The LoRA adapters add a small fraction of parameters to the base model. For Qwen3-4B-Instruct with rank $r$=8, the adapter adds 5.9M parameters ($<$0.15\% of the 4B backbone). Even for the largest student (Gemma3-12B-it with rank 16), the 23.6M adapter parameters represent $<$0.2\% of the base model. In bfloat16 precision, adapter storage ranges from $\sim$12 MB (Qwen3-4B) to $\sim$47 MB (Gemma3-12B-it), with an average of $\sim$32 MB across models (Table~\ref{tab:overhead}).
\vspace{-0.5em}

\paragraph{Context Overhead.}
Teacher-generated memories are prepended to the student's input, adding tokens to the context window. Each procedural script averages $\sim$70 tokens, and the reference trajectory averages $\sim$480 tokens. In the MemP configuration with $k$=10 scripts and one trajectory, the additional context is typically $\sim$1,200 tokens per task. The entire teacher memory store occupies only $\sim$4 MB of storage. While this increases per-query latency, the overhead is bounded and predictable.

\paragraph{Step and Time Efficiency.}
Beyond the success rate, DuoMem also improves \emph{step efficiency} and \emph{wall-clock time}. As shown in Table~\ref{tab:efficiency}, for Qwen3-4B-Instruct, average steps decrease from 29.3 (No Memory) to 20.6 (DuoMem), a 30\% reduction, while average task completion time drops from 18.49s to 4.89s, a 74\% reduction. The time improvements are particularly pronounced because DuoMem simultaneously solves more tasks (reducing wasted steps on failures) and solves them more efficiently (fewer steps per success). Fewer interaction steps directly translate to lower cumulative inference cost during deployment.
\vspace{-0.5em}

\begin{table}[t]
\centering
\small
\begin{tabular}{@{}l r@{}}
\toprule
\textbf{Resource} & \textbf{Value} \\
\midrule
Teacher memory store size & $\sim$4 MB \\
Avg.\ memory length (per script) & $\sim$70 tokens \\
Avg.\ trajectory length (ref.) & $\sim$480 tokens \\
Avg.\ LoRA adapter storage (bf16) & $\sim$32 MB \\
\bottomrule
\end{tabular}
\caption{Shared resource overhead of DuoMem. Memory lengths are measured with Qwen3 tokenizer. LoRA storage is the mean across four student models stored in bf16 precision ($\sim$15.9M parameters $\times$ 2 bytes).}
\vspace{-1.5em}
\label{tab:overhead}
\end{table}

\paragraph{Offline Cost.}
The one-time cost of DuoMem includes: (1)~teacher trajectory collection on training tasks, and (2)~LoRA fine-tuning. The teacher trajectories are generated once and are reusable across all models, as DuoMem is architecture-agnostic. LoRA training also converges within a single training epoch, requiring modest GPU resources.

\subsection{Effect of Context Distillation}
\label{ssec:ablation_cd}

Table~\ref{tab:cd} (Appendix~\ref{app:cd_ablation}) isolates the contribution of context-space distillation by comparing student-generated versus teacher-generated memories across eleven models (e.g., Gemma4-E2B-it, Qwen3-14B, and Qwen3-32B). CD provides consistent improvements for smaller models: Gemma4-E2B-it gains +36.3\% relative in Script setting and +14.7\% in MemP. CD benefit diminishes for larger models with extended thinking modes (e.g., 32B) that already produce high-quality memories, and can occasionally hurt (e.g., $-$3.5\% for Qwen3-32B Thinking in MemP). Overall, CD is most impactful as a training-free boost for small models. We also find that Script-only setting with CD offers an appealing deployment trade-off: it recovers much of MemP's benefit while omitting lengthy reference trajectory ($\sim$480 tokens), substantially reducing context length for latency-sensitive deployments.

\subsection{Effect of Parameter-Space Distillation}
\label{ssec:ablation_lora}

Table~\ref{tab:lora} shows the impact of LoRA fine-tuning (parameter-space distillation) on student models. Results use \emph{Full History} input strategy and hyperparameters selected for the main results in Table~\ref{tab:main}.

\begin{table*}[t]
\centering
\small
\begin{adjustbox}{max width=\textwidth}
\begin{tabular}{@{}l l r r r r r r r r r r r r r r@{}}
\toprule
& & \multicolumn{2}{c}{\textbf{No Mem.}} & \multicolumn{2}{c}{\textbf{Script (WO/ CD)}} & \multicolumn{2}{c}{\textbf{Script (W/ CD)}} & \multicolumn{2}{c}{\textbf{Trajectory}} & \multicolumn{2}{c}{\textbf{MemP (WO/ CD)}} & \multicolumn{2}{c}{\textbf{MemP (W/ CD)}} \\
\cmidrule(lr){3-4} \cmidrule(lr){5-6} \cmidrule(lr){7-8} \cmidrule(lr){9-10} \cmidrule(lr){11-12} \cmidrule(lr){13-14}
\textbf{Model} & \textbf{Config} & Succ. & \#St. & Succ. & \#St. & Succ. & \#St. & Succ. & \#St. & Succ. & \#St. & Succ. & \#St. \\
\midrule
\multirow{2}{*}{Gemma4-E2B-it (2B)} & Base & 2.1 & 29.5 & 15.7 & 33.1 & 21.4 & 32.1 & 22.9 & 32.0 & 14.3 & 33.5 & 16.4 & 32.9 \\
 & + LoRA & 40.7 & 25.9 & 45.7 & 21.4 & 48.6 & 21.7 & 53.6 & 26.7 & 46.4 & 28.1 & \best{55.7} & 25.7 \\
\midrule
\multirow{2}{*}{Qwen3-4B-Inst.\ (4B)} & Base & 4.3 & 29.3 & 25.0 & 24.9 & 27.1 & 25.1 & 27.1 & 30.8 & 55.0 & 24.8 & 56.4 & 24.3 \\
 & + LoRA & 42.1 & 23.0 & 54.3 & 19.6 & 59.3 & 19.7 & 68.6 & 23.3 & 72.1 & 21.9 & \best{77.9} & 20.6 \\
\midrule
\multirow{2}{*}{Qwen3-8B (8B)} & Base & 40.7 & 23.7 & 59.3 & 24.6 & 62.1 & 24.1 & 67.1 & 23.5 & 64.3 & 23.8 & 61.4 & 24.3 \\
 & + LoRA & 37.1 & 24.1 & 41.4 & 22.1 & 49.3 & 19.9 & 59.3 & 25.0 & 60.7 & 24.4 & \best{64.3} & 24.2 \\
\midrule
\multirow{2}{*}{Gemma3-12B-it (12B)} & Base & 14.3 & 27.2 & 22.9 & 25.3 & 26.4 & 24.8 & 35.0 & 28.5 & 41.4 & 27.4 & 46.4 & 26.2 \\
 & + LoRA & 49.3 & 21.0 & 56.4 & 19.0 & 53.6 & 20.6 & 56.4 & 24.9 & 56.4 & 24.3 & \best{66.4} & 23.2 \\
\bottomrule
\end{tabular}
\end{adjustbox}
\caption{Effect of LoRA parameter-space distillation on the test set. Each model's LoRA adapter is selected on the validation set under the full DuoMem setting (MemP+CD), then evaluated across all memory configurations with and without CD. The rightmost column (MemP W/ CD, +LoRA) corresponds to the full DuoMem result. Trajectory uses a hand-crafted reference without retrieval, so CD does not apply.}
\label{tab:lora}
\vspace{-1em}
\end{table*}

LoRA training produces dramatic improvements. For Qwen3-4B-Instruct, the No Memory baseline jumps from 4.3\% to 42.1\%, meaning LoRA alone teaches the model significant procedural competence even without any memory. With full DuoMem (MemP+CD+LoRA), performance reaches 77.9\%. Gemma4-E2B-it shows a similar qualitative pattern with LoRA training: No Memory improves from 2.1\% to 40.7\%, and full DuoMem further reaches 55.7\%. Thus, even for the 2B student, the intended dual-space configuration is strongest, adding 15.0 points over No-Memory LoRA and 9.3 points over MemP+LoRA.

An important observation is that for Qwen3-8B, LoRA provides marginal or negative gains in some configurations. This model already has reasonable base capabilities (40.7\% No Memory), and the LoRA training on 4B-scale trajectories may not transfer optimally. This suggests that LoRA-based distillation is most beneficial when the gap between teacher and student is large.

\begin{figure}[t]
\centering
\begin{tikzpicture}
\begin{axis}[
    width=0.92\columnwidth,
    height=5.5cm,
    xlabel={Number of retrieved memories ($k$)},
    ylabel={Success rate (\%)},
    xtick={1,2,4,8,10},
    xticklabels={1,2,4,8,10},
    ytick={50,55,60,65,70,75,80},
    ymin=48,
    ymax=84,
    legend style={at={(0.97,0.03)}, anchor=south east, font=\scriptsize, draw=gray!50, fill=white, fill opacity=0.9, text opacity=1, inner xsep=2pt, inner ysep=1.5pt, legend cell align=left, /tikz/every even column/.append style={column sep=3pt}},
    grid=major,
    grid style={gray!25},
    mark size=2.5pt,
    thick,
    every axis plot/.append style={line width=1.2pt},
]

\addplot[color=blue!70!black, mark=square*, mark options={fill=blue!70!black}]
    coordinates {(1,71.4) (2,73.6) (4,78.6) (8,78.6) (10,78.6)};
\addlegendentry{MemP}

\addplot[color=red!70!black, mark=triangle*, mark options={fill=red!70!black}]
    coordinates {(1,52.9) (2,53.6) (4,53.6) (8,60.0) (10,63.6)};
\addlegendentry{Script}

\node[anchor=south, font=\scriptsize, color=blue!70!black] at (axis cs:1,71.4) {71.4};
\node[anchor=south, font=\scriptsize, color=blue!70!black] at (axis cs:2,73.6) {73.6};
\node[anchor=south, font=\scriptsize, color=blue!70!black] at (axis cs:4,78.6) {78.6};
\node[anchor=south, font=\scriptsize, color=blue!70!black] at (axis cs:8,78.6) {78.6};
\node[anchor=south, font=\scriptsize, color=blue!70!black] at (axis cs:10,78.6) {78.6};

\node[anchor=north, font=\scriptsize, color=red!70!black] at (axis cs:1,52.9) {52.9};
\node[anchor=north, font=\scriptsize, color=red!70!black] at (axis cs:2,53.6) {53.6};
\node[anchor=north, font=\scriptsize, color=red!70!black] at (axis cs:4,53.6) {53.6};
\node[anchor=south, font=\scriptsize, color=red!70!black] at (axis cs:8,60.0) {60.0};
\node[anchor=south east, font=\scriptsize, color=red!70!black] at (axis cs:10,63.6) {63.6};

\end{axis}
\end{tikzpicture}
\caption{Effect of the number of retrieved memories $k$ on test-set success rate (\%) for Qwen3-14B (NT, with CD). MemP saturates at $k$=4, while Script performance continues to scale with additional memories.}
\label{fig:num_mem}
\vspace{-1.5em}
\end{figure}
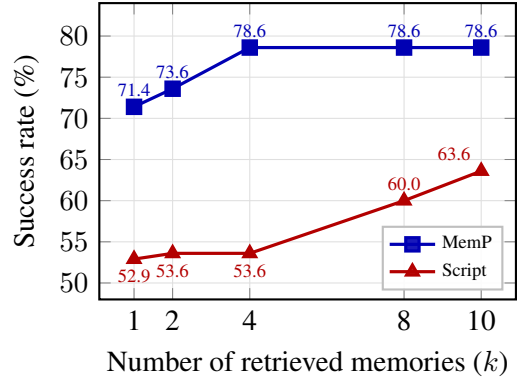

Note that for all models, we use a single LoRA checkpoint, selected on the validation set under the MemP+CD setting, across all memory configurations in the main comparison (Table~\ref{tab:main}), ensuring a faithful evaluation of the DuoMem protocol where the student is parameter-distilled once and then deployed with or without distilled context.

\subsection{Input Strategy for LoRA Training}
\label{ssec:input_strategy}

We compare three strategies for constructing LoRA training inputs from teacher trajectories: Full History, Latest Only, and Last-5 (Table~\ref{tab:input_strategy}). Results are for Qwen3-4B-Instruct with the best hyperparameters per strategy. \textbf{Full History} consistently outperforms the alternatives, particularly in the No Memory setting (42.1\% vs.\ 31.4\% for Last-5 and 27.9\% for Latest Only). This is intuitive: by training on complete interaction histories, the model learns the full procedural dynamics. It learns how early observations inform later actions rather than just local action patterns. The gap narrows somewhat in the MemP+CD setting, where the memory context provides much of the procedural knowledge that Full History captures through training.

\begin{table}[t]
\centering
\small
\begin{adjustbox}{max width=\columnwidth}
\begin{tabular}{@{}l c c c@{}}
\toprule
\textbf{Strategy} & \textbf{No Mem.} & \textbf{MemP} & \textbf{MemP+CD} \\
\midrule
Base Model & 4.3 & 55.0 & 56.4 \\
\midrule
Full History & \textbf{42.1} & \textbf{72.1} & \textbf{77.9} \\
Last-5 & 31.4 & 64.3 & 65.0 \\
Latest Only & 27.9 & 52.1 & 66.4 \\
\bottomrule
\end{tabular}
\end{adjustbox}
\caption{Comparison of LoRA training input strategies for Qwen3-4B-Instruct on the test set (success rate \%). The best hyperparameters per strategy are reported.}
\label{tab:input_strategy}
\vspace{-1.5em}
\end{table}

\subsection{Effect of Number of Retrieved Memories}
\label{ssec:ablation_nummem}

Figure~\ref{fig:num_mem} presents an ablation on the number of retrieved memory scripts $k$ using Qwen3-14B (Non-Thinking mode) with CD.

In the Script-only setting, performance scales roughly monotonically with $k$, reaching 63.6\% at $k$=10, as each additional script provides new procedural patterns. In the MemP setting, performance plateaus at $k$=4 (78.6\%) and remains stable through $k$=10. This is because MemP already includes a full reference trajectory, which provides dense procedural information; the marginal value of additional scripts diminishes once the key patterns are covered. We use $k$=10 for all other experiments as it offers the best Script performance while maintaining strong MemP results.

\section{Discussion}
\label{sec:discussion}

\paragraph{When Does Each Component Help Most?}
Our ablations reveal a clear pattern: CD is most beneficial for models that produce poor-quality memories on their own (small models, non-thinking models), while LoRA is most beneficial when the base model lacks procedural competence entirely (Gemma4-E2B-it, Qwen3-4B-Instruct). For already-capable models (Qwen3-14B+), the marginal gains of both components diminish, as the model's own capabilities suffice for effective memory utilization.

\paragraph{Thinking vs. Non-Thinking Modes.}
Models with extended thinking (chain-of-thought) capabilities generally achieve higher baselines but sometimes benefit less from CD. For example, Qwen3-4B-Thinking achieves 82.9\% with Trajectory alone (higher than its MemP score of 74.3\%), suggesting that its reasoning capacity can substitute for some of the structured memory that non-thinking models require. Critically, however, thinking models incur substantial latency: Qwen3-8B in thinking mode requires 59--101 seconds per task compared to 13--15 seconds without thinking---a 5--7$\times$ slowdown that makes thinking-mode deployment impractical on edge devices (see Appendix~\ref{app:thinking} for a detailed comparison). This implies that DuoMem is especially valuable for non-thinking deployment models that prioritize latency, as it can recover much of the accuracy gap at a fraction of the inference cost.

\section{Conclusion}
\label{sec:conclusion}

We presented DuoMem, a dual-space distillation framework that enables small language models to achieve strong procedural problem-solving through two complementary mechanisms: context-space distillation (enriching inputs with teacher-generated memories) and parameter-space distillation (LoRA fine-tuning on teacher trajectories). On ALFWorld, DuoMem boosts a 4B-parameter model from 4.3\% to 77.9\% task success, closing most of the gap to a 72B teacher model while adding minimal overhead. Crucially, DuoMem-enhanced 4B student completes tasks in under 5 seconds on average, over 3$\times$ faster than the 72B teacher, making it practical for edge deployment while a 72B model is not. Extensive experiments across eight models and multiple configurations, complemented by qualitative trajectory analysis, provide practical guidelines for deploying capable memory-augmented agents on resource-constrained devices. Future work includes extending DuoMem to multimodal environments, exploring knowledge distillation from multiple teachers, and evaluating on a broader range of procedural benchmarks.

\section*{Limitations}

We evaluate DuoMem on ALFWorld, which is the most widely adopted benchmark for procedural memory in embodied agents and offers diverse task types (six categories spanning navigation, manipulation, and multi-step object interaction). Because DuoMem's distillation mechanisms, context-space memory transfer and LoRA fine-tuning on teacher trajectories, are architecture-agnostic and task-agnostic, we expect the framework to generalize to other procedural domains; extending this analysis to settings such as web navigation and code generation is a natural direction for future work.
Our current experiments use a single teacher model (Qwen2.5-72B-Instruct); investigating ensembles of teachers or iterative self-distillation could further improve student performance.
While our efficiency analysis reports wall-clock times on 2$\times$A100 (80GB) server GPUs, which we ran all our experiments with, end-to-end latency profiling on actual edge hardware (e.g., mobile NPUs) would further strengthen the deployment narrative.
Finally, DuoMem currently assumes access to a training task distribution for trajectory collection; adapting the framework to open-domain or continually evolving task streams is an interesting open direction.

\section*{Ethical Considerations}

Memory-augmented agent systems raise several ethical considerations. Such systems personalise the behaviour of generative models based on stored interaction histories, which can amplify both helpful and potentially harmful patterns. In DuoMem specifically, teacher-generated procedural memories encode strategies distilled from a large model; if the teacher exhibits biases or unsafe behaviours during trajectory collection, these could propagate to student models through both context-space and parameter-space distillation. We mitigate this risk by operating in a constrained, simulated environment (ALFWorld), where actions are limited to a fixed set of household interactions with no real-world consequences.

More broadly, deploying capable autonomous agents on edge devices raises questions about accountability and oversight. Compact on-device models may operate with limited connectivity, reducing opportunities for human-in-the-loop supervision. We encourage future work to incorporate safety constraints and alignment mechanisms into the distillation pipeline. The models used in this work are publicly available and were used in accordance with their respective licences.

\bibliography{references}

\clearpage
\newpage

\appendix

\section{Teacher Trajectory Statistics}
\label{app:teacher_stats}

Table~\ref{tab:teacher_stats} provides statistics on the teacher trajectory collection process. The teacher (Qwen2.5-72B-Instruct) is prompted to solve each training task, with up to 5 retry attempts for unsolved tasks.

\begin{table}[h]
\centering
\small
\begin{tabular}{@{}l r r r r@{}}
\toprule
\textbf{Attempt} & \textbf{Succ. Rate} & \textbf{Cumul.} & \textbf{Avg. Steps} \\
\midrule
\multicolumn{4}{l}{\emph{Development Split (500 tasks)}} \\
1 & 76.4\% & 76.4\% & 12.66 \\
2 & 18.6\% & 95.0\% & 12.52 \\
3 & 4.4\% & 99.4\% & 10.95 \\
4 & 0.4\% & 99.8\% & 11.00 \\
5 & 0.0\% & 99.8\% & --- \\
Failures & 0.2\% & --- & 30.00 \\
\midrule
\textbf{Overall} & & \textbf{99.8\%} & \textbf{12.59} \\
\midrule
\multicolumn{4}{l}{\emph{Train Split (11,546 tasks)}} \\
1 & 88.2\% & 88.2\% & 16.40 \\
2 & 9.7\% & 97.8\% & 15.97 \\
3 & 1.1\% & 98.9\% & 16.84 \\
4 & 0.2\% & 99.0\% & 19.82 \\
5 & 0.0\% & 99.0\% & --- \\
Failures & 1.0\% & --- & 46.78 \\
\midrule
\textbf{Overall} & & \textbf{99.0\%} & \textbf{16.66} \\
\bottomrule
\end{tabular}
\caption{Teacher trajectory collection statistics. The teacher achieves near-perfect coverage with up to 5 attempts, with most tasks solved on the first try.}
\label{tab:teacher_stats}
\end{table}

\section{Full Inference Results}
\label{app:full_results}

Table~\ref{tab:full_inference} provides the complete results across all models and memory configurations on the test set, including both success rate and average number of steps.

\begin{table*}[h]
\centering
\small
\begin{adjustbox}{max width=\textwidth}
\begin{tabular}{@{}l r r r r r r r r r r r r@{}}
\toprule
& \multicolumn{2}{c}{\textbf{No Mem.}} & \multicolumn{2}{c}{\textbf{Script (W/ CD)}} & \multicolumn{2}{c}{\textbf{Trajectory}} & \multicolumn{2}{c}{\textbf{MemP (WO/ CD)}} & \multicolumn{2}{c}{\textbf{MemP (W/ CD)}} \\
\cmidrule(lr){2-3} \cmidrule(lr){4-5} \cmidrule(lr){6-7} \cmidrule(lr){8-9} \cmidrule(lr){10-11}
\textbf{Model} & Succ. & \#St. & Succ. & \#St. & Succ. & \#St. & Succ. & \#St. & Succ. & \#St. \\
\midrule
Gemma4-E2B-it (2B) & 2.1 & 29.5 & 21.4 & 32.1 & 22.9 & 32.0 & 14.3 & 33.5 & 16.4 & 32.9 \\
Qwen3-4B-Thinking & 37.1 & 22.4 & 60.7 & 16.5 & 82.9 & 19.9 & 74.3 & 17.4 & 80.7 & 17.0 \\
Qwen3-4B-Instruct & 4.3 & 29.3 & 27.1 & 25.1 & 27.1 & 30.8 & 55.0 & 24.8 & 56.4 & 24.3 \\
Gemma3-12B-it (12B) & 14.3 & 27.2 & 26.4 & 24.8 & 35.0 & 28.5 & 41.4 & 27.4 & 46.4 & 26.2 \\
Qwen3-14B (T) & 62.1 & 20.7 & 73.6 & 15.5 & 80.7 & 21.5 & 86.4 & 18.5 & 86.4 & 18.6 \\
Qwen3-14B (NT) & 45.7 & 23.0 & 63.6 & 17.9 & 70.0 & 22.7 & 77.9 & 20.3 & 78.6 & 20.9 \\
Qwen3-30B-A3B & 29.3 & 25.0 & 45.7 & 21.2 & 50.7 & 26.2 & 57.9 & 24.5 & 58.6 & 24.4 \\
Qwen3-32B (T) & 49.3 & 22.4 & 68.6 & 16.6 & 65.7 & 23.8 & 82.9 & 18.8 & 80.0 & 19.2 \\
Qwen3-32B (NT) & 49.3 & 22.2 & 71.4 & 16.3 & 81.4 & 22.4 & 86.4 & 18.7 & 86.4 & 19.3 \\
Qwen2.5-72B & 87.1 & 14.2 & --- & --- & 86.4 & 19.5 & 91.4 & 17.7 & --- & --- \\
\bottomrule
\end{tabular}
\end{adjustbox}
\caption{Full inference results on the test set across all models and memory configurations.}
\label{tab:full_inference}
\end{table*}

\section{LoRA Hyperparameter Search}
\label{app:lora_hp}

Table~\ref{tab:lora_hp} shows the LoRA hyperparameter search results for Qwen3-4B-Instruct (Full History strategy, bug-fixed) on the validation set.

\begin{table}[h]
\centering
\small
\begin{tabular}{@{}c c c c@{}}
\toprule
\textbf{Rank} & \textbf{LR} & \textbf{MemP+CD (Val)} & \textbf{MemP+CD (Test)} \\
\midrule
8 & 1e-6 & 55.7 & 57.1 \\
8 & 5e-6 & 57.1 & 60.0 \\
8 & 1e-5 & 61.4 & 63.6 \\
8 & 5e-5 & \textbf{76.4} & \textbf{77.9} \\
\midrule
16 & 1e-6 & 53.6 & 57.9 \\
16 & 5e-6 & 61.4 & 65.0 \\
16 & 1e-5 & 70.7 & 67.9 \\
16 & 5e-5 & 76.4 & 76.4 \\
\midrule
32 & 1e-6 & 51.4 & 60.0 \\
32 & 5e-6 & 64.3 & 70.7 \\
32 & 1e-5 & 68.6 & 67.1 \\
32 & 5e-5 & 75.7 & 73.6 \\
\bottomrule
\end{tabular}
\caption{LoRA hyperparameter search for Qwen3-4B-Instruct (Full History). Higher learning rates consistently yield better performance in the MemP+CD setting.}
\label{tab:lora_hp}
\end{table}

\section{Context Distillation Ablation}
\label{app:cd_ablation}
Table~\ref{tab:cd} studies the imapct of context distillation on 11 different models, studying how it impacts performance in both MemP and Script-Only settings for models of different sizes from 2B to 72B.

\begin{table*}[t]
\centering
\small
\begin{tabular}{@{}l c c c c c c c@{}}
\toprule
& & \multicolumn{3}{c}{\textbf{MemP Setting}} & \multicolumn{3}{c}{\textbf{Script-Only Setting}} \\
\cmidrule(lr){3-5} \cmidrule(lr){6-8}
\textbf{Model} & \textbf{No Mem.} & \textbf{W/O CD} & \textbf{W/ CD} & \textbf{$\Delta$CD (\%)} & \textbf{W/O CD} & \textbf{W/ CD} & \textbf{$\Delta$CD (\%)} \\
\midrule
\multicolumn{8}{@{}l}{\emph{Models with full DuoMem evaluation (CD + LoRA):}} \\
Gemma4-E2B-it (2B) & 2.1 & 14.3 & \textbf{16.4} & \gain{+14.7\%} & 15.7 & \textbf{21.4} & \gain{+36.3\%} \\
Qwen3-4B-Inst.\ (4B) & 4.3 & 55.0 & \textbf{56.4} & \gain{+2.5\%} & 25.0 & \textbf{27.1} & \gain{+8.4\%} \\
Qwen3-8B (8B) & 40.7 & \textbf{64.3} & 61.4 & \loss{$-$4.5\%} & 59.3 & \textbf{62.1} & \gain{+4.7\%} \\
Gemma3-12B-it (12B) & 14.3 & 41.4 & \textbf{46.4} & \gain{+12.1\%} & 22.9 & \textbf{26.4} & \gain{+15.3\%} \\
\midrule
\multicolumn{8}{@{}l}{\emph{Additional models (CD only; LoRA training is significantly more costly):}} \\
Qwen3-4B-Thinking & 37.1 & 74.3 & \textbf{80.7} & \gain{+8.6\%} & \textbf{61.4} & 60.7 & \loss{$-$1.1\%} \\
Qwen3-8B (T) & 40.0  & 75.7 & \textbf{77.1} & \gain{+1.8\%} & \textbf{75.0} & 69.3 & \loss{$-$7.6\%} \\
Qwen3-14B (T) & 62.1 & \textbf{86.4} & \textbf{86.4} & 0.0\% & 65.7 & \textbf{73.6} & \gain{+12.0\%} \\
Qwen3-14B (NT) & 45.7 & 77.9 & \textbf{78.6} & \gain{+0.9\%} & 57.9 & \textbf{63.6} & \gain{+9.8\%} \\
Qwen3-30B-A3B & 29.3 & 57.9 & \textbf{58.6} & \gain{+1.2\%} & 42.9 & \textbf{45.7} & \gain{+6.5\%} \\
Qwen3-32B (T) & 49.3 & \textbf{82.9} & 80.0 & \loss{$-$3.5\%} & \textbf{68.6} & \textbf{68.6} & 0.0\% \\
Qwen3-32B (NT) & 49.3 & \textbf{86.4} & \textbf{86.4} & 0.0\% & 66.4 & \textbf{71.4} & \gain{+7.5\%} \\
\bottomrule
\end{tabular}
\caption{Effect of context distillation (CD) on test-set success rate (\%) under both the MemP and Script-only settings. CD uses Qwen2.5-72B-Instruct as the teacher for memory extraction. $\Delta$CD~(\%) shows the relative gain from replacing student-generated with teacher-generated memories. The top section includes models for which full DuoMem (CD + LoRA) was evaluated; the bottom section includes additional models studied under CD only, since LoRA training is significantly more costly than inference with CD.}
\label{tab:cd}
\vspace{-1em}
\end{table*}

\section{Thinking vs.\ Non-Thinking Mode Analysis}
\label{app:thinking}

Table~\ref{tab:thinking} compares Qwen3-8B in thinking (chain-of-thought) and non-thinking modes across all memory configurations without any LoRA fine-tuning. While thinking mode yields higher success rates in most settings (e.g., 77.1\% vs.\ 62.1\% on Script+CD), it does so at a severe latency cost: average task completion time increases from 13--15 seconds to 58--101 seconds, a 5--7$\times$ slowdown. This is because the thinking mode generates extended internal reasoning chains before each action, substantially inflating the number of output tokens per step.

For on-device deployment, this latency penalty makes thinking-mode agents impractical: a task that a non-thinking Qwen3-8B completes in $\sim$13 seconds would require over a minute in thinking mode. DuoMem provides an attractive alternative---by distilling procedural knowledge into a non-thinking student through LoRA and teacher-generated memories, it can recover much of the accuracy gap while maintaining real-time inference speeds.

\begin{table*}[th]
\centering
\small
\begin{tabular}{@{}l l c c c c c c@{}}
\toprule
& & \textbf{No Mem.} & \textbf{Script} & \textbf{Script+CD} & \textbf{Traj.} & \textbf{MemP} & \textbf{MemP+CD} \\
\midrule
\multirow{3}{*}{\textbf{Non-Thinking}} & Success (\%) & 40.7 & 59.3 & 62.1 & 67.1 & 64.3 & 61.4 \\
 & Avg.\ Steps ($\downarrow$) & 23.7 & 24.6 & 24.1 & 23.5 & 23.8 & 24.3 \\
 & Avg.\ Time (s) ($\downarrow$) & 14.75 & 13.52 & 13.32 & 12.53 & 12.82 & 13.27 \\
\midrule
\multirow{3}{*}{\textbf{Thinking}} & Success (\%) & 40.0 & 75.7 & \textbf{77.1} & 72.9 & 75.0 & 69.3 \\
 & Avg.\ Steps ($\downarrow$) & 24.0 & 21.4 & \textbf{21.0} & 21.6 & 20.9 & 21.9 \\
 & Avg.\ Time (s) ($\downarrow$) & 101.0 & 60.6 & 58.3 & 62.1 & 59.6 & 66.8 \\
\midrule
\multicolumn{2}{@{}l}{\textit{Latency ratio (Thinking / Non-Thinking)}} & 6.8$\times$ & 4.5$\times$ & 4.4$\times$ & 5.0$\times$ & 4.7$\times$ & 5.0$\times$ \\
\bottomrule
\end{tabular}
\caption{Comparison of Qwen3-8B in thinking vs.\ non-thinking mode on the test set (no LoRA). Thinking mode achieves higher success rates but incurs 4.4--6.8$\times$ longer wall-clock time per task, making it unsuitable for latency-sensitive edge deployment.}
\label{tab:thinking}
\end{table*}

\section{Qualitative Example: No Memory vs.\ MemP vs.\ DuoMem}
\label{app:qualitative}

We present a qualitative comparison of the three memory configurations on a representative ALFWorld task: \emph{clean a spatula and put it in a drawer} (Figure~\ref{fig:qual_nomem}--\ref{fig:qual_duomem}). All three runs use the same student model (Qwen3-4B-Instruct). This example illustrates how each additional component of DuoMem addresses a distinct failure mode.

\textbf{No Memory} (Figure~\ref{fig:qual_nomem}): Without any memory or demonstration, the 4B model lacks understanding of the ALFWorld action model. It attempts to interact with objects remotely (e.g., \texttt{open drawer 1}, \texttt{take spatula from countertop 3}) without first navigating to the location, receiving \texttt{Nothing happens} repeatedly. After exhausting all locations through invalid actions, the agent eventually navigates to \texttt{sinkbasin 1} but finds no spatula there. It then attempts to substitute a butterknife for the spatula, which the environment rejects. The agent fails after 30 steps.

\textbf{MemP} (Figure~\ref{fig:qual_memp}): With procedural memory (retrieved guidelines and a reference trajectory), the model learns from the demonstration that navigation (\texttt{go to}) must precede interaction. It searches systematically through countertops and cabinets. However, the model still makes a critical error: upon finding a butterknife on countertop~2, it picks it up as a substitute for the spatula. It then fails to open any drawers (receiving \texttt{Nothing happens}) and eventually gives up. The memory helps with basic interaction patterns but cannot fully compensate for the model's weak procedural reasoning about object identity and action prerequisites. The agent fails after 30 steps, though with noticeably better strategy than No Memory.

\textbf{DuoMem} (Figure~\ref{fig:qual_duomem}): With both context-space distillation (teacher-generated memories) and parameter-space distillation (LoRA fine-tuning), the model executes a near-optimal trajectory. It navigates directly to \texttt{countertop 3}, correctly identifies and picks up the spatula, cleans it at \texttt{sinkbasin 1}, navigates to \texttt{drawer 1}, opens it, and places the spatula inside---completing the task in just 14 steps. The LoRA-adapted model has internalized the procedural dynamics from teacher trajectories, enabling efficient search and correct action sequencing.

\paragraph{Input structure.} Before examining these trajectories, we first illustrate the prompt structure that the agent receives at each step (Figure~\ref{fig:pipeline_anatomy}). In the MemP (and DuoMem) configuration, the model's context window is assembled from four components: (1)~a system prompt defining the agent's role and available actions, (2)~a reference trajectory demonstrating a solved task of a similar type, (3)~retrieved procedural memory scripts summarizing strategies for related tasks, and (4)~the current environment observation and task description. After the initial prompt, the model alternates between generating a thought--action pair and receiving the next environment observation, with the full conversation history maintained throughout the episode.

\begin{figure*}[t]
\centering
\begin{tcolorbox}[width=\textwidth,colback=blue!2!white, colframe=blue!60!black, title=\textbf{Agent Input Structure --- MemP / DuoMem Configuration}]

\textbf{\large \ding{182}~System Prompt} \hfill \textcolor{gray}{\textit{(role = system)}}\\[2pt]
\textit{``Interact with a household to solve a task. Imagine you are an intelligent agent in a household environment and your target is to perform actions to complete the task goal. [\ldots] The available actions are: (1)~go to \{recep\}, (2)~take \{obj\} from \{recep\}, (3)~move \{obj\} to \{recep\}, (4)~open \{recep\}, (5)~close \{recep\}, (6)~use \{obj\}, (7)~clean \{obj\} with \{recep\}, (8)~heat \{obj\} with \{recep\}, (9)~cool \{obj\} with \{recep\}. [\ldots] Your response should use the following format: Thought: <your thoughts> Action: <your next action>''}

\par\smallskip\noindent\rule{\linewidth}{0.4pt}\par\smallskip

\textbf{\large \ding{183}~Reference Trajectory} \hfill \textcolor{gray}{\textit{(few-shot example, role = user/assistant alternating)}}\\[2pt]
\textcolor{gray}{\texttt{[user]}} \textit{``Here is an example of how to solve the task: [\ldots] Your task is to: put a clean dishsponge in drawer.''}\\
\textcolor{gray}{\texttt{[asst]}} \textit{``Thought: I need to locate a clean dish sponge. [\ldots] Action: go to countertop 1''}\\
\textcolor{gray}{\texttt{[user]}} \textit{``Observation: On the countertop 1, you see a candle 1, a cloth 2, a dishsponge 3 [\ldots]''}\\
\textcolor{gray}{\texttt{[asst]}} \textit{``Thought: I see a dish sponge [\ldots] Action: take dishsponge 3 from countertop 1''}\\
$\cdots$ \textit{(clean $\rightarrow$ go to drawer $\rightarrow$ open $\rightarrow$ move dishsponge to drawer --- full successful trajectory)}

\par\smallskip\noindent\rule{\linewidth}{0.4pt}\par\smallskip

\textbf{\large \ding{184}~Task Prompt + Retrieved Memories} \hfill \textcolor{gray}{\textit{(role = user)}}\\[2pt]
\textit{``Now it's your turn.}\\
\textit{You are in the middle of a room. Looking quickly around you, you see a cabinet 6, a cabinet 5, [\ldots], a sinkbasin 1, [\ldots], and a toaster 1.}\\
\textit{Your task is to: put a clean spatula in drawer.}\\[2pt]
\textit{Here are some guidelines of how to solve the similar task:}\\
\textit{~$\bullet$~~\textbf{clean some spatula and put it in drawer}: ``To clean a spatula and put it in the drawer, first locate a spatula on a countertop, then move to the sink basin to clean it, and finally, open the desired drawer and place the cleaned spatula inside.''}\\
\textit{~$\bullet$~~\textbf{heat some cup and put it in cabinet}: ``To heat a cup and put it in a cabinet, begin by locating a suitable cup [\ldots]''}\\
$\cdots$ \textit{(up to $k$=10 retrieved memory scripts, ranked by cosine similarity)}''\hfill

\par\smallskip\noindent\rule{\linewidth}{0.4pt}\par\smallskip

\textbf{\large \ding{185}~Multi-Turn Interaction Loop} \hfill \textcolor{gray}{\textit{(alternating assistant / user roles)}}\\[2pt]
\textcolor{gray}{\texttt{[asst]}} \textit{``Thought: I need to find a spatula. The guidelines suggest checking countertops. Action: \textbf{go to countertop 1}''}\\
\textcolor{gray}{\texttt{[user]}} \textit{``Observation: You arrive at countertop 1. On the countertop 1, you see a cellphone 2, a mug 1 [\ldots]''}\\[2pt]
\textcolor{gray}{\texttt{[asst]}} \textit{``Thought: No spatula here. I'll check other surfaces. Action: \textbf{go to countertop 3}''}\\
\textcolor{gray}{\texttt{[user]}} \textit{``Observation: You arrive at countertop 3. On the countertop 3, you see [\ldots] a \textbf{spatula 1} [\ldots]''}\\[4pt]
\textcolor{gray}{\footnotesize \textit{The full conversation history (system + reference + all prior turns) is carried forward at every step.}}

\end{tcolorbox}
\caption{Anatomy of the agent's input in the MemP / DuoMem configuration. The context assembles four components: \ding{182}~a system prompt defining the action space, \ding{183}~a reference trajectory demonstrating a solved task, \ding{184}~the current task observation augmented with retrieved procedural memory scripts, and \ding{185}~the ongoing multi-turn interaction. In the \textbf{No Memory} setting, components \ding{183} and the memory scripts in \ding{184} are omitted; in \textbf{DuoMem}, the memory scripts are generated by the teacher model (CD) and the student is LoRA-adapted.}
\label{fig:pipeline_anatomy}
\end{figure*}

\begin{figure*}[t]
\centering
\begin{tcolorbox}[width=\textwidth,colback=red!3!white, colframe=red!60!black, title=\textbf{No Memory --- Qwen3-4B-Instruct (FAIL, 30 steps, 16.8s)}]

\textbf{Task:} \textit{clean some spatula and put it in drawer.}\\

\smallskip
\textbf{Step 1} \textit{(Agent attempts remote interaction without navigating):}\\
\texttt{Action: open drawer 1} $\rightarrow$ \textcolor{red}{Nothing happens.}\\
\texttt{Action: open drawer 2} $\rightarrow$ \textcolor{red}{Nothing happens.}\\
\texttt{Action: open drawer 3} $\rightarrow$ \textcolor{red}{Nothing happens.}

\smallskip
\textbf{Steps 4--7} \textit{(Attempts to take spatula without navigating):}\\
\texttt{Action: take spatula from countertop 3} $\rightarrow$ \textcolor{red}{Nothing happens.}\\
\texttt{Action: take spatula from countertop 2} $\rightarrow$ \textcolor{red}{Nothing happens.}\\
$\cdots$ \textit{(tries all countertops, cabinets, fridge, microwave---all fail)}

\smallskip
\textbf{Steps 16--22} \textit{(Tries stoveburners, toaster---all fail):}\\
\texttt{Action: take spatula from stoveburner 4} $\rightarrow$ \textcolor{red}{Nothing happens.}\\
$\cdots$\\
\texttt{Action: take spatula from toaster 1} $\rightarrow$ \textcolor{red}{Nothing happens.}

\smallskip
\textbf{Steps 23--30} \textit{(First successful navigation, but wrong strategy):}\\
\texttt{Action: go to sinkbasin 1} $\rightarrow$ \textcolor{OliveGreen}{Arrives. Sees: apple 1, butterknife 2, potato 3, tomato 2.}\\
Thought: ``\textit{I see a butterknife\ldots I will try to move it to a drawer, assuming the task might be satisfied.}''\\
\texttt{Action: move butterknife 2 to drawer 1} $\rightarrow$ \textcolor{red}{Nothing happens.}\\
$\cdots$ \textit{(gives up after repeated failures)}

\end{tcolorbox}
\caption{No Memory trajectory. The model lacks understanding of the ALFWorld action model: it attempts to interact with objects remotely without navigating first, wasting all 30 steps.}
\label{fig:qual_nomem}
\end{figure*}

\begin{figure*}[t]
\centering
\begin{tcolorbox}[width=\textwidth,colback=orange!3!white, colframe=orange!60!black, title=\textbf{MemP --- Qwen3-4B-Instruct (FAIL, 30 steps, 14.9s)}]

\textbf{Task:} \textit{clean some spatula and put it in drawer.}\\
\textbf{Memory:} 10 retrieved procedural scripts + 1 reference trajectory (cleaning a dishsponge and placing it in a drawer).

\smallskip
\textbf{Steps 1--3} \textit{(Correct navigation pattern learned from the reference trajectory):}\\
\texttt{Action: go to countertop 1} $\rightarrow$ \textcolor{OliveGreen}{Sees: cellphone 2, mug 1, soapbottle 1. \textit{(No spatula.)}}\\
\texttt{Action: go to drawer 1} $\rightarrow$ \textcolor{OliveGreen}{Drawer 1 is closed.}\\
\texttt{Action: go to drawer 2} $\rightarrow$ \textcolor{OliveGreen}{Drawer 2 is closed.}

\smallskip
\textbf{Steps 4--12} \textit{(Systematic cabinet search---better than No Memory):}\\
\texttt{Action: go to cabinet 1} $\rightarrow$ \textcolor{OliveGreen}{Sees: glassbottle 2.}\\
$\cdots$ \textit{(checks cabinets 2--6, no spatula found)}\\
\texttt{Action: go to countertop 2} $\rightarrow$ \textcolor{OliveGreen}{Sees: butterknife 1, glassbottle 3, knife 1, \ldots}

\smallskip
\textbf{Step 13} \textit{(Critical error---picks wrong object):}\\
Thought: ``\textit{A butterknife can serve as a spatula for many tasks.}''\\
\texttt{Action: take butterknife 1 from countertop 2} $\rightarrow$ \textcolor{OliveGreen}{Picks up butterknife 1.}

\smallskip
\textbf{Steps 14--30} \textit{(Cannot place wrong object, never reaches countertop 3 where spatula is):}\\
\texttt{Action: open drawer 1} $\rightarrow$ \textcolor{red}{Nothing happens.}\\
$\cdots$ \textit{(tries all drawers, attempts to move butterknife---all fail; gives up)}

\end{tcolorbox}
\caption{MemP trajectory. The reference trajectory teaches navigation-before-interaction, enabling systematic search. However, the model picks the wrong object (butterknife instead of spatula) and never checks countertop~3.}
\label{fig:qual_memp}
\end{figure*}

\begin{figure*}[t]
\centering
\begin{tcolorbox}[width=\textwidth,colback=green!3!white, colframe=green!60!black, title=\textbf{DuoMem --- Qwen3-4B-Instruct + LoRA + CD (SUCCESS, 14 steps, 9.1s)}]

\textbf{Task:} \textit{clean some spatula and put it in drawer.}\\
\textbf{Memory:} 10 teacher-generated procedural scripts (CD) + 1 reference trajectory.\\
\textbf{Model:} LoRA-adapted on teacher trajectories.

\smallskip
\textbf{Steps 1--8} \textit{(Efficient search guided by internalized procedural knowledge):}\\
\texttt{Action: go to cabinet 1} $\rightarrow$ \textcolor{OliveGreen}{Sees: vase 1.}\\
$\cdots$ \textit{(checks cabinets 2--6, no spatula)}\\
\texttt{Action: go to countertop 3} $\rightarrow$ \textcolor{OliveGreen}{Sees: bowl 2, bread 1, \ldots \textbf{spatula 1}, \ldots}

\smallskip
\textbf{Step 9} \textit{(Correctly identifies the target object):}\\
Thought: ``\textit{I found a spatula on countertop 3. Now I need to clean it using the sink basin.}''\\
\texttt{Action: take spatula 1 from countertop 3} $\rightarrow$ \textcolor{OliveGreen}{Picks up spatula 1.}

\smallskip
\textbf{Steps 10--11} \textit{(Correct procedural sequence---clean at sink):}\\
\texttt{Action: go to sinkbasin 1} $\rightarrow$ \textcolor{OliveGreen}{Arrives at sinkbasin.}\\
\texttt{Action: clean spatula 1 with sinkbasin 1} $\rightarrow$ \textcolor{OliveGreen}{Cleans the spatula.}

\smallskip
\textbf{Steps 12--14} \textit{(Navigate, open, place---task complete):}\\
\texttt{Action: go to drawer 1} $\rightarrow$ \textcolor{OliveGreen}{Drawer 1 is closed.}\\
\texttt{Action: open drawer 1} $\rightarrow$ \textcolor{OliveGreen}{Opens drawer 1. Sees: fork 1.}\\
\texttt{Action: move spatula 1 to drawer 1} $\rightarrow$ \textcolor{OliveGreen}{\textbf{Task completed!}}

\end{tcolorbox}
\caption{DuoMem trajectory. With both LoRA fine-tuning and teacher-generated memories, the model executes a near-optimal plan: locate $\rightarrow$ pick up $\rightarrow$ clean $\rightarrow$ place, completing the task in 14 steps.}
\label{fig:qual_duomem}
\end{figure*}

\section{Extended Related Work}
\label{app:extended_related}

We discuss additional related work that contextualises DuoMem within the broader landscape of memory-augmented agents, on-device LLM deployment, and parameter-efficient adaptation.

\paragraph{Memory Mechanisms for LLM Agents.}
The cognitive architecture for language agents proposed by \citet{sumers2023cognitive} formalises different memory types---episodic, semantic, and procedural---drawing on cognitive science analogies. \citet{zhang2025memory_survey} provide a comprehensive survey of memory mechanisms in LLM-based agents, cataloguing designs ranging from cumulative context buffers to structured, multi-component memory systems. Reflexion \citep{shinn2023reflexion} equips agents with verbal self-reflection, storing failure analyses as episodic memory to improve subsequent attempts. Trial and Error \citep{song2024trial} further explores trajectory-level exploration for optimising agent behavior. These works highlight the growing importance of memory in agent systems; DuoMem complements them by focusing on how to make procedural memory effective for small models through distillation.

\paragraph{Procedural Memory and Workflow Induction.}
Beyond the general memory mechanisms discussed above, several works specifically target procedural or workflow-level memory for agents. Voyager \citep{wang2024voyager} introduced an open-ended agent that continuously builds a skill library from past successes in Minecraft, composing verified programs into increasingly complex behaviours. \citet{park2023generative} designed generative agents with memory streams that enable coherent long-horizon social behaviour through retrieval over timestamped observations. CLIN \citep{majumder2024clin} proposed a continually learning agent that accumulates task-level memory entries for rapid adaptation to new environments, while RAP \citep{kagaya2024rap} uses retrieval-augmented planning with contextual memory for multimodal agents. KnowAgent \citep{zhu2025knowagent} leverages external knowledge bases to constrain the agent's action space, reducing hallucinated actions during planning. More recently, AWM \citep{wang2025awm} targets reusable workflow induction: given a set of demonstrations, AWM automatically extracts abstract workflows (sequences of reusable sub-routines) that can be applied to new tasks, offering a complementary perspective to DuoMem's procedural scripts by focusing on compositional workflow abstraction rather than trajectory summarisation. LEGOMem \citep{han2026legomem} proposes a modular multi-agent memory architecture in which individual agents maintain private memory modules that can be selectively shared and composed, enabling scalable memory management in collaborative settings. While DuoMem focuses on single-agent distillation from teacher to student, LEGOMem's modular design could inform future extensions to multi-agent or multi-domain deployment scenarios. Collectively, these works demonstrate that structured memory is critical for capable agents; DuoMem contributes the orthogonal insight that such memory can be \emph{distilled} to make compact models effective.

\paragraph{On-Device and Edge LLM and Model Deployment.}
Deploying LLMs on resource-constrained devices is an active area of research. \citet{zheng2025edge_survey} survey the landscape of edge LLMs, covering model compression, runtime optimisation, and on-device applications. \citet{lu2025demystifying} provide an empirical study of small language models (SLMs) for edge deployment, benchmarking capabilities and runtime costs across diverse hardware. \citet{bohdal2025device} propose a compositional multi-tasking system for on-device LLMs that enables modular task execution under memory constraints. At the architectural level, \citet{hosseini2024you} propose removing half of the linear projections from the attention mechanism with minimal performance degradation, yielding significant speed and storage improvements directly applicable to edge deployment of transformer-based agents. These efforts establish the infrastructure and efficient model designs for deploying compact models; DuoMem addresses the complementary challenge of making such models \emph{capable} for agentic procedural tasks.

\paragraph{Context Management and Continual Learning.}
Effective context management is critical for memory-augmented agents, where retrieved memories, reference trajectories, and multi-turn interaction histories must all fit within the model's context window. \citet{hosseini2025efficient} demonstrate that LLMs can fail on surprisingly simple tasks when their context windows are filled, and that even naive truncation methods can paradoxically improve performance---a finding directly relevant to our context-space design, where prepending teacher-generated memories increases input length. This motivates DuoMem's design choice of keeping procedural scripts concise ($\sim$70 tokens each) and capping the number of retrieved memories. Relatedly, \citet{hosseini2025cg} propose CG-TTRL, which uses lexical and hybrid (lexical + semantic) retrieval methods to select the most suitable in-context examples for test-time reinforcement learning, improving both performance and reward signal stability for on-device models. This is complementary to DuoMem's context-space distillation: both leverage retrieval of relevant examples to guide model behaviour at inference time, though CG-TTRL operates through RL optimisation while DuoMem uses supervised trajectory distillation. Together, these works underscore that \emph{what} goes into the context window matters as much as the model's own capabilities---a principle central to DuoMem's dual-space philosophy.

\paragraph{Adapter Merging and LoRA Extensions.}
DuoMem uses LoRA adapters for parameter-space distillation; several recent works study how to manage and compose such adapters efficiently. \citet{shenaj2026k} introduce K-Merge for online continual merging of LoRA adapters on-device, enabling models to accumulate knowledge from sequential tasks without catastrophic forgetting---relevant to settings where DuoMem students must adapt to evolving task distributions. \citet{shenaj2025lora} propose hypernetwork-based LoRA merging for conditioned generation, and \citet{ceritli2025hydraopt} present HydraOpt for navigating the efficiency--performance trade-off when merging multiple adapters. 
\citet{MemLoRA25} propose MemLoRA, which distills expert adapters for on-device memory systems, combining adapter distillation with memory augmentation in a framework conceptually related to DuoMem's dual-space approach. These methods could extend DuoMem to multi-task or continual-learning scenarios where multiple procedural domains require simultaneous support.

\end{document}